# PT-DETR: Small Target Detection Based on Partially-Aware Detail Focus

Bingcong Huo, Zhiming Wang

**ABSTRACT** To address the challenges in UAV object detection, such as complex backgrounds, severe occlusion, dense small objects, and varying lighting conditions, this paper proposes PT-DETR based on RT-DETR, a novel detection algorithm specifically designed for small objects in UAV imagery. In the backbone network, we introduce the Partially-Aware Detail Focus (PADF) Module to enhance feature extraction for small objects. Additionally, we design the Median-Frequency Feature Fusion (MFFF) module, which effectively improves the model's ability to capture small-object details and contextual information. Furthermore, we incorporate Focaler-SIoU to strengthen the model's bounding box matching capability and increase its sensitivity to small-object features, thereby further enhancing detection accuracy and robustness. Compared with RT-DETR, our PT-DETR achieves mAP improvements of 1.6% and 1.7% on the VisDrone2019 dataset with lower computational complexity and fewer parameters, demonstrating its robustness and feasibility for small-object detection tasks.

I. INTRODUCTION

Unmanned aerial vehicles (UAVs) are widely used in military reconnaissance[1], agricultural monitoring, forest fire prevention, disaster relief, and traffic management due to their low cost, ease of operation, and adaptability to various environments. The aerial images captured by UAVs offer wide coverage, flexible viewpoints, high information density, and rich content. However, due to the unique characteristics of UAV imaging and the complexity of typical aerial scenes[2], challenges such as a low proportion of small targets, dense target overlap, and occlusion by terrain often arise, making accurate object detection highly challenging. Small targets[3] are generally defined as objects with a resolution smaller than 32×32 pixels. Their features are difficult to extract because of the small object size and low pixel count, and in complex scenes, dense small targets are easily affected by noise and background interference. Additionally, in images of extremely small targets, the distribution of foreground and background information is highly imbalanced, resulting in suboptimal performance for standard detection models under such conditions. This poses a new research direction for object detection, as the accurate identification and localization of small targets demand higher precision and robustness from detection algorithms.

With the rapid development of deep learning, it has become a core technology for UAV-based[4] object detection. Current mainstream detection algorithms are mainly divided into two categories: convolutional neural network (CNN)-based architectures and Transformer-based architectures. Lim et al. [5]proposed a method that leverages contextual connections across multi-scale features by using additional features from different network depths as context and combining them with an attention mechanism to focus on objects in the image, thereby fully exploiting contextual information and improving small-object detection accuracy. Simonyan et al. [6] proposed using multiple small convolutional kernels instead of large ones to achieve multi-scale object detection, enhancing small-target detection performance while maintaining low computational cost. Zhang Liangliang et al. [7] adjusted the detection head size and network structure, introduced the Biformer attention mechanism, and designed a DAT detection head to improve feature fusion, though this approach incurs higher computational costs. Overall, CNN-based detection algorithms still face a trade-off between accuracy and efficiency when handling dense or overlapping small targets.

In 2024, Baidu proposed the RT-DETR[8] model, which demonstrated excellent performance in both accuracy and speed. RT-DETR quickly showed strong adaptability across various tasks, and researchers began applying it to small-object detection. J Hu et al. [9] designed a context-guided feature fusion module to enhance the model's ability to merge multi-scale features while preventing the loss of fine-grained information. Renzheng Xue et al. [10] introduced an attention-based intra-scale feature interaction that incorporates an effective additional feature selection mechanism, thereby reducing the computational complexity of the model.

To address the challenges of object detection in aerial images, this paper proposes an efficient UAV image detection Transformer framework, named PT-DETR. We introduce the PADF Module, which combines the redundant computation compression capability of PConv with the cross-scale dynamic focusing property of the PAT partial attention mechanism. In addition, we design the Multi-Scale Feature Refinement Pyramid, whose core components — the SPDConv module and the Median-Frequency Feature Fusion (MFFF) module — effectively enhance the model's ability to capture small-object details and contextual information. Finally, we replace the original GIoU with Focaler-SIoU to strengthen the model's bounding box matching capability and improve its sensitivity to small-object features, thereby further enhancing detection accuracy and robustness for small targets.

## II. RELATED WORK

### A. RT-DETR

RT-DETR (Real-Time Detection Transformer) is a novel real-time object detector proposed by the Baidu team in 2024. Its core breakthrough lies in constructing, for the first time, an end-to-end real-time object detection framework that completely eliminates the reliance on post-processing techniques such as Non-Maximum Suppression (NMS).

The main innovations of RT-DETR include the hybrid encoder, which serves as the core feature processing unit and contains two key submodules: Adaptive Interaction Feature Integration (AIFI) module, which focuses on processing high-level features output by the backbone network. Cross-scale Feature Fusion Module (CCFM), responsible for integrating and interacting multi-scale features to strengthen the relationships between features at different scales.

The decoder is equipped with an auxiliary prediction head and iteratively refines object queries to generate the final bounding boxes and confidence scores. Notably, the initial queries for the decoder are not generated randomly; instead, they are dynamically adjusted by the encoder based on IoU. The encoder selects image features from the output sequence that pay the most attention to the detection target regions, which are then used as the initial queries to improve subsequent prediction accuracy.

B. Feature Fusion Method

The core of feature fusion lies in effectively combining features from different sources and levels to generate a more comprehensive and robust feature representation, providing a stronger foundation for subsequent object detection tasks. The development in this field began with the concept of pyramid features[11], which led to various efficient fusion solutions. For example, Lin et al. [12] proposed the Feature Pyramid Network (FPN), which was the first to effectively aggregate high-resolution low-level features with low-resolution high-level features, laying the foundation for cross-scale feature fusion.

Subsequently, a series of optimized frameworks emerged and were applied to object detection, all demonstrating excellent performance: Liu S et al.[13] proposed PANet : Enhances bidirectional interaction between low-level and high-level features to improve the capture of small-object features. NAS-FPN [14]: Uses Neural Architecture Search (NAS) to automatically search for optimal feature fusion paths, reducing manual design effort. ASFF [15]: Employs an adaptive spatial feature fusion mechanism to dynamically adjust the fusion weights of features at different scales. BiFPN [16]: Simplifies the feature fusion path and adds cross-scale connections, improving performance while reducing computational cost.

To further address issues such as information inconsistency and insufficient feature representation in feature fusion, researchers have proposed optimizations from different perspectives: AugFPN [17] introduces one-time supervision during the information fusion stage to specifically address inconsistencies between detailed and semantic information in feature maps, narrowing the gap between different types of information. Cheng et al. [18] use dual attention modules before feature fusion to enhance features and guide the network to focus on key dimensions of the target. Liu et al. [19] proposed the High-Resolution Detection Network (HRDNet), which combines multi-depth image pyramids with multi-scale FPNs to deepen the feature representation hierarchy, thereby improving the perception of small objects.

III. METHODOLOGY

The proposed PT-DETR algorithm is built upon the RT-DETR architecture. Its Transformer-based design effectively models global context and long-range dependencies, which is crucial for detecting small objects in complex UAV imagery. RT-DETR is chosen because its end-to-end detection pipeline eliminates the need for NMS post-processing, thereby reducing inference latency in real-time applications. Furthermore, the lightweight RT-DETR-R18 variant provides an optimal balance between performance and computational efficiency, making it suitable for UAV platforms.

Compared to DETR, our algorithm introduces three improvements. First, in the backbone network, we design a lightweight Partially-Aware Detail Focus (PADF) module. By combining Partial Convolution (PConv)[20] with the PTA attention mechanism[21], this enhancement reduces computational redundancy through PConv while promoting multi-scale feature fusion and cross-spatial interaction via PTA, enabling efficient modeling of both global and local features.

Second, in the neck network, we propose the Multi-Scale Feature Refinement Pyramid (MSFRP), which includes the SPDConv module[22] and the Median-Frequency Feature Fusion Module (MFFF). These modules leverage SPDConv to extract small-object-relevant features and fuse multi-scale feature maps, while global median pooling and depthwise channel and frequency-domain attention mechanisms enhance inter-channel interactions and improve feature representation.

Finally, we replace the original GIoU[23] with Focaler-SIoU to strengthen the model's bounding box matching capability and improve its sensitivity to small-object features, thereby further enhancing detection accuracy and robustness for small targets.

The overall framework of PT-DETR is illustrated in Fig.1.

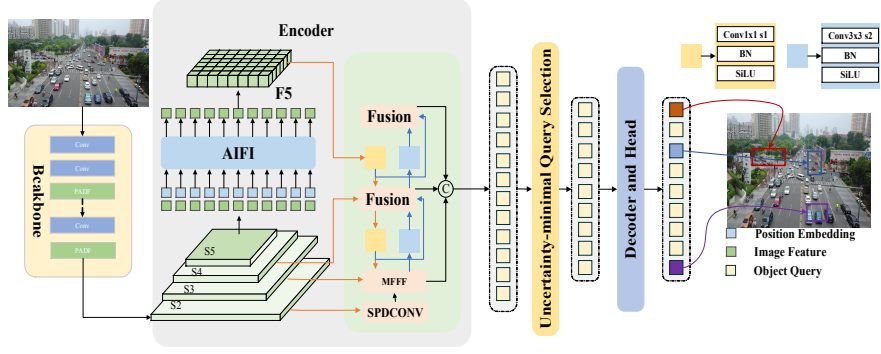

Fig.1. The diagram of PT-DETR model structure.

A. PADF Module

To improve feature extraction efficiency, this study integrates Partial Convolution (PConv) with the PTA attention mechanism to design a multi-scale, efficient, and fast feature extraction module, named PADF Module (Partially-Aware Detail Focus module), as shown in Fig.2(a). Currently, to optimize performance, feature extraction commonly employs Depthwise Convolution (DWConv)[24] or Group Convolution (GConv)[25] . However, although these operations reduce FLOPs, they increase memory access, creating a speed bottleneck. Partial Convolution (PConv) divides the input feature channels into two parts, performing convolution only on the first C_conv channels while leaving the remaining channels unchanged, thereby significantly reducing computation and the number of parameters. Compared with standard convolution, PConv lowers FLOPs while retaining essential feature information, enabling higher inference efficiency without sacrificing accuracy.

The PTA attention mechanism consists of two components: the partial channel attention module (PAT_ch), which combines 3×3 convolutions with channel attention to achieve global spatial interaction, as illustrated in Fig.2(b); and the partial spatial attention module (PAT_sp), which integrates 1×1 convolutions with spatial attention to efficiently fuse channel information, as shown in Fig.2(c).

The PADF module innovatively integrates PConv with the PTA partial attention mechanism to construct an efficient feature extraction architecture. By leveraging PConv to reduce computational redundancy and combining it with PTA to promote multi-scale feature fusion and cross-spatial interactions, it enables efficient modeling of both global and local features. Through the collaborative optimization of local perception and global dependencies, this module can significantly enhance feature representation in complex scenarios.

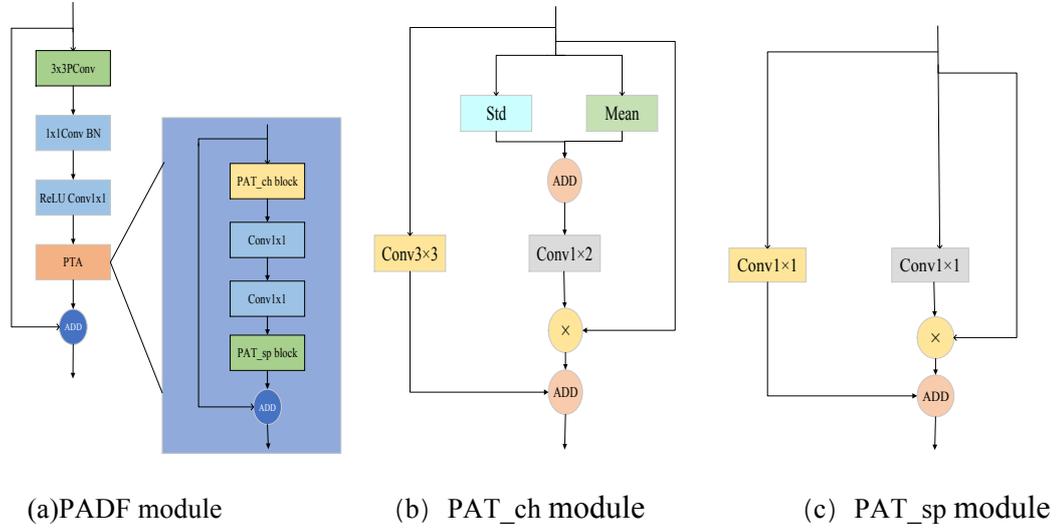

(a) PADF module    (b) PAT_ch module    (c) PAT_sp module

Fig.2. The diagram of PADF module structure.

B. Multi-Scale Feature Refinement Pyramid

Although RT-DETR achieves relatively promising performance in object detection, its neck network still relies on a traditional pyramid structure, which cannot fully leverage the correlations between different feature maps, resulting in suboptimal feature fusion. In addition, UAV aerial images often exhibit highly imbalanced sample distributions and frequent severe occlusions, and the fixed-size convolutional receptive fields in the neck network are insufficient to capture fine-grained local information.

To address these issues, this paper proposes a Multi-Scale Feature Refinement Pyramid (MSFRP) module. Unlike the conventional approach of simply adding a P2 detection layer, we process the P2 feature map through a downsampling module (SPDConv) to extract features rich in small-object information and pass them to the P3 layer for fusion. Subsequently, we introduce a Median-Frequency Feature Fusion (MFFF) module, which combines channel attention and frequency-domain attention mechanisms to enhance feature extraction effectiveness and robustness. Specifically, channel attention extracts global statistical information via global pooling, while frequency-domain attention processes feature maps in the frequency domain to improve feature extraction efficiency.

1) SPDCONV

The fine-grained details of small objects tend to become blurred or even lost, making it difficult for the model to accurately capture these subtle yet critical features. To address the challenges of small object detection, detection models first extract features at different scales through the backbone network and then perform feature fusion, combining the rich details from high-resolution feature maps with the semantic information from low-resolution feature maps. This allows the model to both focus on the fine features of small objects and accurately identify them within the overall contextual environment, thereby improving detection accuracy.

In RT-DETR, cross-scale feature interaction still relies on FPN, which is a relatively optimal choice from a real-time perspective. The CCFF module performs PAFPN operations on the S3 to S5 layers, with its fusion module constructed in a CSPBlock style[26]. However, performing cross-scale fusion solely on the S3 to S5 layers has certain limitations. High-resolution feature maps in the multi-scale pyramid, such as S1 and S2, contain abundant fine-grained information that is crucial for small object detection. Relying only on S3 to S5 layers prevents full utilization of these high-resolution features. Incorporating the S2 layer, however, introduces new challenges, the most notable of which is the increased computational cost.

To address these issues in aerial imagery, SPDConv (Spatially Separated and Deformable Convolution) is applied to process the S2 features, reducing the resolution of the S2 layer through spatial downsampling. In the spatial dimensions of the input feature map, SPDConv samples four different spatial levels, enabling it to handle various spatial structures and details within the feature map. The four sampled outputs are then concatenated along the channel dimension to form a new feature map, making the number of channels in the new feature map four times that of the original. The way SPDConv extracts samples at four different spatial locations is shown in Equation (1):

$$x_1 = x[..,::2,::2]$$
$$x_2 = x[..,1::2,::2]$$

$$x_3=x[..,::2,1::2]$$
$$x_4=x[..,1::2,1::2] \quad (1)$$

The key to the sampling is to capture information from different spatial locations, i.e., selecting features from different regions along the height (H) and width (W) dimensions, which helps address the varying spatial structures and details in the feature map. Then, the four sampled features are concatenated along the channel dimension to form a new feature map. That is, the number of channels in the resulting feature map is four times that of the original channels.

The SPDConv module concatenates information from different spatial regions to better enhance its spatial awareness. The concatenated feature map is then processed by a standard convolution. Finally, the SPDConv module outputs the resulting convolutional feature map. The structure of the SPDConv module is shown in Fig.3.

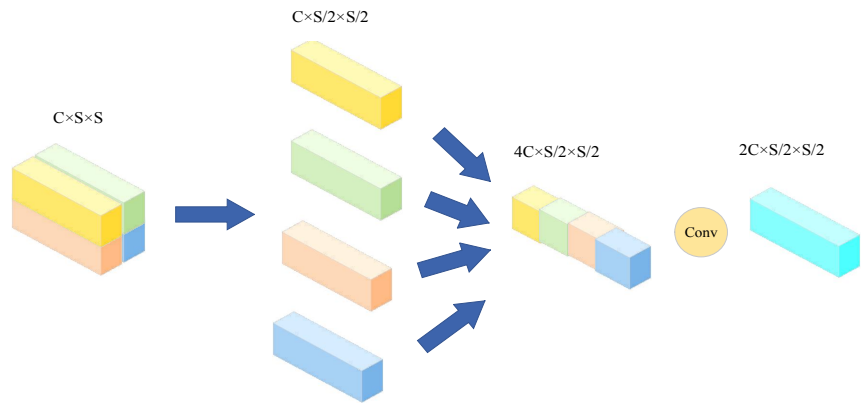

Fig.3. The diagram of SPDConv model structure.

2) Median-Frequency Feature Fusion Module (MFFF)

In UAV small object detection tasks, due to the small target size and the susceptibility of feature information to loss, designing an efficient and robust feature extraction module becomes a key factor in improving detection performance. To this end, we design and propose a novel module—the Median-Frequency Feature Fusion Module (MFFF), as shown in Figure 4. The MFFF module effectively enhances feature representation by combining channel attention with frequency-domain attention mechanisms, thereby improving the model's performance in small object detection scenarios.

The design of the MFFF module incorporates two core innovations. First, one branch introduces a Global Median Pooling (GMP) operation to capture more comprehensive global feature statistics. Second, the other branch employs a deep channel attention mechanism along with a frequency-domain attention mechanism to enhance feature extraction capabilities. Through these two designs, the MFFF module significantly improves the model's detection accuracy while incurring minimal computational overhead.

In the first branch, in addition to the commonly used Global Average Pooling (GAP) and Global Max Pooling (GMP), the Global Median Pooling (GMP) operation is introduced as a complement. Global Median Pooling computes the median of the feature map along the spatial dimensions, providing a more robust representation of the feature distribution. The formula is as follows:

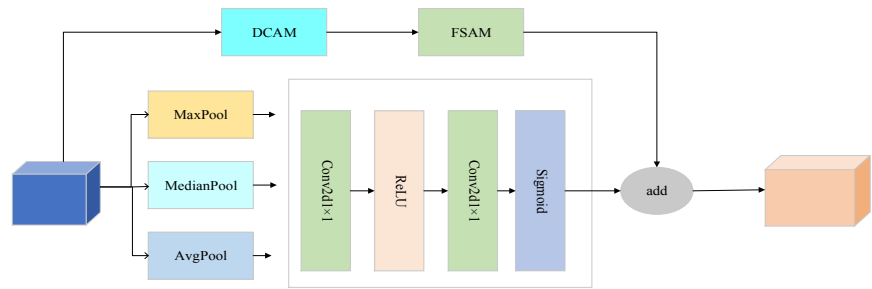

(a) MFFF module

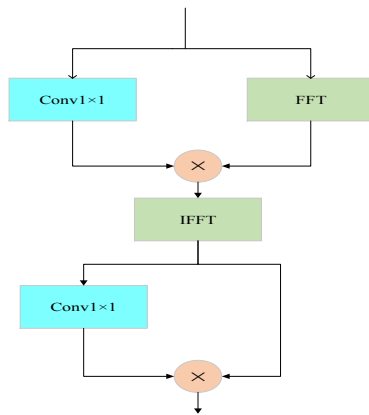

(b) DCAM module

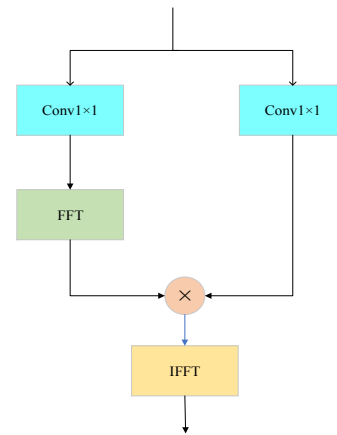

(c) FSAM module

Fig.4. The diagram of MFFF model structure

In the other branch, the deep channel attention mechanism uses a dual-channel attention approach to enhance mutual information between feature channels, thereby improving their representation capability. First, the input features undergo a 1×1 convolution and are then transformed into the frequency domain via the Fast Fourier Transform (FFT). In this domain, dual-channel weighting is applied through two distinct convolution paths, processing feature information across different channels. The features are then transformed back to the spatial domain using the Inverse Fast Fourier Transform (IFFT) and passed through another 1×1 convolution to reconstruct the feature map. Finally, these features are fused to further strengthen inter-channel correlations and enhance feature expressiveness.

The frequency-domain attention mechanism enhances feature extraction efficiency by processing the feature map in the frequency domain. First, the input features undergo a 1×1 convolution and are transformed to the frequency domain via FFT. In this domain, weighted attention emphasizes different frequency components to enhance critical frequency information. The feature map is then transformed back to the spatial domain through IFFT. Finally, the attention-enhanced frequency-domain features are fused with the original features to obtain a richer feature representation.

C. Focaler-SIoU Loss Function

In object detection tasks, the accuracy of bounding box regression plays a crucial role in overall detection performance. Localization errors not only reduce detection precision but also increase false positives and false negatives, thereby weakening the model's robustness in complex scenarios. This issue is particularly pronounced in UAV aerial imagery, where severe sample imbalance exists: a small number of hard-to-detect targets (such as pedestrians and bicycles) contribute the majority of gradients during training, while easily detectable samples are often ignored. As a result, traditional IoU-based losses are limited in their ability to achieve high-precision localization.

To address this problem, this paper introduces the Focaler-SIoU loss function in the regression branch, enhancing and improving upon traditional IoU-based losses.

Focaler-IoU [27] focuses on the distribution of hard and easy samples, allowing the loss function to be adjusted according to the specific requirements of different detection tasks, thereby improving the model's efficiency in handling various types of samples. Additionally, the use of linear interval mapping further strengthens the

attention to diverse regression samples. The corresponding calculation formula is shown in Equation 2:

$$IoU^{focaler}=\begin{cases} 0, & IoU<<d \\ \dfrac{IoU-d}{u-d}, & d<<IoU<<u \\ 1, & IoU>>u \end{cases} \quad (2)$$

Here, IoU represents the original intersection-over-union value, while d and u are dynamic thresholds that adjust the loss computation across different IoU intervals. By adjusting the parameters u and d, this method controls the sensitivity of the loss function to different samples, enabling more precise localization on high-quality samples while suppressing interference from low-quality samples. [d, u] ϵ [0, 1] The corresponding loss is defined in Equation 3:

$$L_{Focaler-Iou}=1-IoU^{focaler} \quad (3)$$

When applied to SIoU-based[28] bounding box regression, Focaler-SIoU enables the model to focus more on highly overlapping and well-matched samples, thereby achieving finer bounding box adjustments, as shown in Equation 4:

$$L_{Focaler-SIou}=L_{SIoU}+IoU-IoU^{focaler} \quad (4)$$

Compared with traditional IoU-based loss functions (such as GIoU, DIoU, CIoU, and EIoU), Focaler-SIoU not only considers the geometric overlap between predicted and ground-truth boxes but also integrates an adaptive weighting strategy based on sample difficulty. This improvement allows the model to dynamically adjust its focus during optimization, effectively enhancing the accuracy of bounding box regression and the overall detection performance.

IV. EXPERIMENTS

A. Experimental Setup

In this experiment, the VisDrone2019 dataset [29] is used. As shown in Fig. 5, the dataset contains 6,471 images in the training set, 548 images in the validation set, and 1,580 images in the test set. The annotation categories cover a wide variety of scenes from UAV perspectives, including 10 different object classes: pedestrian, person, bicycle, car, van, truck, tricycle, awning-tricycle, bus, and motorbike.

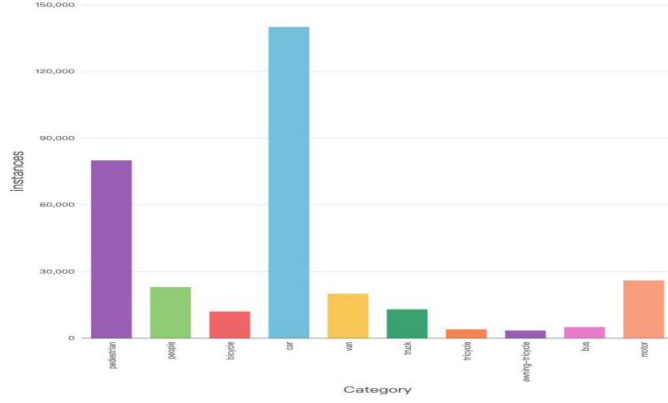

Fig.5. Target category distributuin diagram

B. Experimental Environment

As shown in Table 1, the experimental hardware environment consists of a Tesla P40 GPU with 24 GB of memory and an Intel(R)Xeon(R)CPU E5-2678 v3 @ 2.50GHZ. The deep learning framework used is PyTorch 2.1.0, and the operating system is Ubuntu 20.04. To ensure the reproducibility of the experiments and the comparability of the ablation studies, the hyper parameters are set as follows: the total number of training epochs is 300, the batch size is 4, and the input image size is 640 × 640. The Adam[30] optimizer is used with an initial learning rate of 0.0001, the momentum parameter is 0.9, and the weight decay coefficient is 0.0001.

Table.1. Experimental parameter settings

| Parameter | Value |
| :---: | :---: |
| Epochs | 300 |
| Batch Size | 4 |
| Imgsz | 640×640 |
| Optimizer | AdamW |
| Weigh_decay | 0.0001 |
| Monmentum | 0.9 |

C. Evaluate Metrics

To accurately evaluate the improvement of the proposed PT-DETR, we use mean Average Precision (mAP), model parameters (Params), giga floating-point operations per second (GFLOPs) as metrics for assessing model performance, detailed as follows.

Precision (P) represents the proportion of correctly predicted samples among all samples predicted as positive. It is calculated using Equation 5, where TP denotes the number of correctly predicted objects and FP denotes the number of incorrectly predicted objects.

$$P = \frac{TP}{TP+FP} \quad (5)$$

Recall (R) represents the proportion of actual positive samples that are correctly predicted. It is calculated using Equation 6, where FN denotes the number of existing objects that were not correctly detected.

$$R = \frac{TP}{TP+FN} \quad (6)$$

Average Precision (AP) measures the precision at different thresholds along the precision–recall (PR) curve and is calculated using Equation 7. The mean Average Precision (mAP) is the average of AP across all classes, calculated using Equation 9.

$$AP = \int_0^1 P_i(R_i) dR_i \quad (7)$$

$$mAP = \frac{\sum_{i=1}^{N} AP_i}{N} \quad (8)$$

To better reflect the model's performance on objects of different sizes, we evaluate the model using $mAP_{50}$ and $mAP_{50\text{-}95}$ metrics. Specifically, $mAP_{50}$ represents the mAP when the IOU is 0.5, and $mAP_{50\text{-}95}$ represents the average mAP when the IOU ranges from 0.5 to 0.95.

D. Ablation Experiments

To verify the contribution of each module in PT-DETR to small object detection performance, we conducted ablation experiments on the VisDrone2019 dataset under the same experimental environment and training parameters. The experimental results are shown in Table 2.

The baseline model RT-DETR achieved a $mAP_{50}$ of 36.8%, and an $mAP_{50\text{-}95}$ of 26.4%. Model A integrates the PADF module into the backbone, achieving a $mAP_{50}$ of 37.2%, an improvement of 0.4% over the baseline, while also reducing the number of parameters by 0.6%. These results demonstrate that replacing the base block in the backbone with the PADF module enables more effective feature extraction, thereby improving object detection performance.

Model B incorporates the proposed MFRP into the baseline model to enhance multi-scale feature fusion. The results show that adding MFRP significantly increases the $mAP_{50\text{-}95}$ by 1.2%, allowing the model to capture finer details of small targets and obtain richer feature representations, while also improving $mAP_{50}$.

The detection results of Model C indicate that when both the PADF module and Focaler-SIoU are added to the baseline model, the model achieves more accurate localization for high-quality samples and outperforms RT-DETR in all detection metrics, confirming the positive contribution of each module to improving detection performance.

Finally, when using Model D (with all modules combined), PT-DETR achieves the best detection performance, with a $mAP_{50}$ of 38.4%, representing a 1.6% improvement over the baseline, fully demonstrating the superior capability of the proposed method in small object detection tasks.

Table.2. Results of ablation experiments.

| Model | 1 | 2 | 3 | Params | GFLOPs | mAP50 | mAP |
|---|---|---|---|---|---|---|---|
| Baseline | | | | 20.09 | 68 | 36.8 | 26.4 |
| A | √ | | | 19.46 | 65.4 | 37.2 | 27.1 |
| B | | √ | | 20.45 | 72.3 | 37.6 | 27.6 |
| C | √ | √ | | 19.46 | 65.4 | 37.4 | 27.3 |
| D | √ | √ | √ | 19.79 | 67.5 | 38.4 | 28.1 |

E.  Comparison Experiment

To further verify the superiority of the proposed PT-DETR, we conducted a quantitative comparison on the VisDrone2019 dataset against several recent state-of-the-art methods.

As shown in Table 3, compared with the baseline RT-DETR-R18, our PT-DETR improves $mAP_{50}$ by 1.6% and $mAP_{50\text{-}95}$ by 1.7%.

Compared with UAV-DETR, which also adopts the DETR architecture, PT-DETR reduces the model size by 0.3 MB, while achieving improvements of 0.8% and 0.6% in $mAP_{50}$ and $mAP_{50\text{-}95}$, respectively.

Compared with the YOLO series of algorithms, PT-DETR achieves a maximum improvement of 3.8% in $mAP_{50}$ and 3.5% in $mAP_{50\text{-}95}$, fully demonstrating its superior performance and strong generalization capability in small object detection tasks.

Table.3. Results of comparison experiments.

| Model | Params | GFLOPs | $mAP_{50}$ | $mAP_{50-95}$ |
|---|---|---|---|---|
| YOLOv8-M | 25.9 | 79.1 | 34.6 | 24.6 |
| YOLOv11-M | 20.2 | 67.4 | 36.4 | 25.9 |
| YOLOv12-M | 20.1 | 66.8 | 35.3 | 25.5 |
| DETR | 60 | 187 | 35.6 | 24.3 |
| RTDETR-R18 | 20.09 | 77 | 36.8 | 26.4 |
| UAV-DETR | 20.15 | 78 | 37.6 | 27.5 |
| PT-DETR(ours) | 19.79 | 67.5 | 38.4 | 28.1 |

F. Visualization

In Fig.6, we present the detection results of small objects in the VisDrone dataset. Compared with the baseline model, PT-DETR shows a significant improvement in its ability to localize small objects. In the prediction maps of our model, more small object categories are detected, indicating that the model can capture the features of these small objects more effectively. Therefore, as shown in the yellow box in Fig.6(b), PT-DETR also performs well in localizing occluded objects.

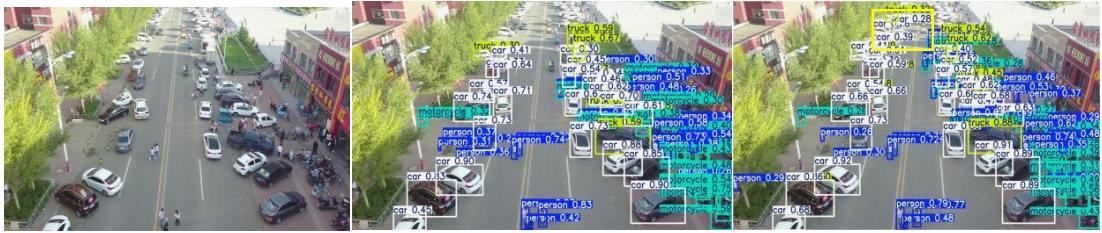

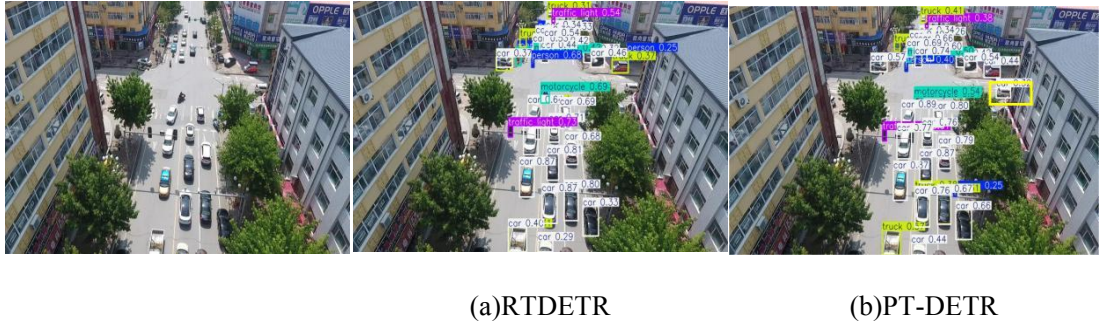

           (a)RTDETR     (b)PT-DETR

Fig.6. Comparison of detection effect

V. CONCLUSION

 We propose PT-DETR, a Transformer-based detection model specifically designed for small object detection. By incorporating the Partially-Aware Detail Focus Block and the Multi-Scale Feature Refinement Pyramid, PT-DETR can effectively extract small object features and fuse spatial and frequency-domain features to better capture high-resolution details, thereby enhancing the expressiveness of feature maps. In addition, we introduce the Focaler-SIoU Loss Function, which allows the model to dynamically adjust its focus during optimization, effectively improving bounding box regression accuracy and overall detection performance.

 Experimental results on the VisDrone-2019-DET dataset demonstrate that PT-DETR achieves competitive detection accuracy compared with existing state-of-the-art methods. Ablation studies further validate the effectiveness of each module, highlighting the model's ability to address the challenges of small object detection within a Transformer-based framework.

 Future work will focus on further improving small object localization and detection accuracy by better balancing high-resolution feature extraction with semantic understanding.

REFERENCES

[1] Mohsan S A H, Khan M A, Noor F, et al. Towards the unmanned aerial vehicles (UAVs): A comprehensive review[J]. Drones, 2022, 6(6): 147.

[2] Z. Yang, J. Lian, and J. Liu, ''Infrared UAV target detection based on continuous-coupled neural network,'' Micromachines, vol. 14, no. 11, p. 2113, Nov. 2023.

[3] Wang X, Wang A, Yi J, et al. Small object detection based on deep learning for remote sensing: A comprehensive review[J]. Remote Sensing, 2023, 15(13): 3265.

[4] Tang G, Ni J, Zhao Y, et al. A survey of object detection for UAVs based on deep learning[J]. Remote Sensing, 2023, 16(1): 149.

[5] Lim J S, Astrid M, Yoon H J, et al. Small object detection using context and attention[C]//2021 international Conference on Artificial intelligence in information and Communication (ICAIIC). IEEE, 2021: 181-186.

[6] Simonyan K, Zisserman A. Very deep convolutional networks for large-scale image recognition[J]. arXiv preprint arXiv:1409.1556, 2014.

[7] Zhu L, Wang X, Ke Z, et al. Biformer: Vision transformer with bi-level routing attention[C]//Proceedings of the IEEE/CVF conference on computer vision and pattern recognition. 2023: 10323-10333.

[8] Zhao Y, Lv W, Xu S, et al. Detrs beat yolos on real-time object detection[C]//Proceedings of the IEEE/CVF conference on computer vision and pattern recognition. 2024: 16965-16974.

[9] Hu J, Liao Z, Bian C, et al. Dynamic Context-guided Feature Fusion Network for gastrointestinal tumor image segmentation[J]. Alexandria Engineering Journal, 2025, 130: 1029-1043.

[10] Xue R, Hua S, Xu H. FECI-RTDETR A lightweight unmanned aerial vehicle infrared small target detector Algorithm based on RT-DETR[J]. IEEE Access, 2025.

[11] Li H. Rethinking Features-Fused-Pyramid-Neck for Object Detection[C]//European Conference on Computer Vision. Cham: Springer Nature Switzerland, 2024: 74-90.

[12] Lin T Y, Dollár P, Girshick R, et al. Feature pyramid networks for object detection[C]//Proceedings of the IEEE conference on computer vision and pattern recognition. 2017: 2117-2125.

[13] Liu S, Qi L, Qin H, et al. Path aggregation network for instance segmentation[C]//Proceedings of the IEEE conference on computer vision and pattern recognition. 2018: 8759-8768.

[14] Ghiasi G, Lin T Y, Le Q V. Nas-fpn: Learning scalable feature pyramid architecture for object detection[C]//Proceedings of the IEEE/CVF conference on computer vision and pattern recognition. 2019: 7036-7045.

[15] Liu S, Huang D, Wang Y. Learning spatial fusion for single-shot object detection. arXiv 2019[J]. arXiv preprint arXiv:1911.09516, 1911.

[16] Tan M, Pang R, Le Q V. Efficientdet: Scalable and efficient object detection[C]//Proceedings of the IEEE/CVF conference on computer vision and pattern recognition. 2020: 10781-10790.

[17] Guo C, Fan B, Zhang Q, et al. Augfpn: Improving multi-scale feature learning for object detection[C]//Proceedings of the IEEE/CVF conference on computer vision and pattern recognition. 2020: 12595-12604.

[18] Cheng G, Lang C, Wu M, et al. Feature enhancement network for object detection in optical remote sensing images[J]. Journal of Remote Sensing, 2021.

[19] Liu Z, Gao G, Sun L, et al. HRDNet: High-resolution detection network for small objects[J]. arXiv preprint arXiv:2006.07607, 2020.

[20] Liu G, Reda F A, Shih K J, et al. Image inpainting for irregular holes using partial convolutions[C]//Proceedings of the European conference on computer vision (ECCV). 2018: 85-100.

[21] Huang H, Xia T, Ren P. Partial Channel Network: Compute Fewer, Perform Better[J]. arXiv preprint arXiv:2502.01303, 2025.

[22] Sunkara R, Luo T. No more strided convolutions or pooling: A new CNN building block for low-resolution images and small objects[C]//Joint European conference on machine learning and knowledge discovery in databases. Cham: Springer Nature Switzerland, 2022: 443-459.

[23] Rezatofighi H, Tsoi N, Gwak J Y, et al. Generalized intersection over union: A metric and a loss for bounding box regression[C]//Proceedings of the IEEE/CVF conference on computer vision and pattern recognition. 2019: 658-666.

[24] Howard A G, Zhu M, Chen B, et al. Mobilenets: Efficient convolutional neural networks for mobile vision applications[J]. arXiv preprint arXiv:1704.04861, 2017.

[25] Xie S, Girshick R, Dollár P, et al. Aggregated residual transformations for deep neural networks[C]//Proceedings of the IEEE conference on computer vision and pattern recognition. 2017: 1492-1500.

[26] Wang C Y, Liao H Y M, Wu Y H, et al. CSPNet: A new backbone that can enhance learning capability of CNN[C]//Proceedings of the IEEE/CVF conference on computer vision and pattern recognition workshops. 2020: 390-391.

[27] Zhang H, Zhang S. Focaler-iou: More focused intersection over union loss[J]. arXiv preprint arXiv:2401.10525, 2024.

[28] Gevorgyan Z. SIoU loss: More powerful learning for bounding box regression[J]. arXiv preprint arXiv:2205.12740, 2022.

[29] P. Zhu, L. Wen, D. Du, X. Bian, H. Fan, Q. Hu, and H. Ling, "Detection and tracking meet drones challenge," IEEE Transactions on Pattern Analysis and Machine Intelligence, vol. 44, no. 11, pp.7380–7399, 2021.

[30] I. Loshchilov, "Decoupled weight decay regularization," arXiv preprint arXiv:1711.05101, 2017.

[31] Wang H, Yu Y, Tang Z. FDM-RTDETR: A Multi-scale Small Target Detection Algorithm[J]. IEEE Access, 2025.

[32] Zhang H, Liu K, Gan Z, et al. UAV-DETR: Efficient End-to-End Object Detection for Unmanned Aerial Vehicle Imagery. arXiv 2025[J]. arXiv preprint arXiv:2501.01855.